\documentclass[final]{cvpr}

\usepackage{times}
\usepackage{epsfig}
\usepackage{graphicx}
\usepackage{amsmath}
\usepackage{amssymb}
\usepackage{subcaption}
\usepackage{bbm}
\usepackage{mathtools}

\usepackage{booktabs}
\usepackage{enumitem}
\usepackage{lipsum}
\usepackage{algorithm,algorithmic}
\usepackage{multicol}
\usepackage{multirow}
\usepackage[pagebackref=true,breaklinks=true,colorlinks,bookmarks=false]{hyperref}

\newtheorem{theorem}{Theorem}
\newtheorem{definition}{Definition}[section]

\def\cvprPaperID{****} 
\def\confYear{CVPR 2021}

\begin{document}

\title{Disentangling Label Distribution for Long-tailed Visual Recognition}

\author{Youngkyu Hong\thanks{Equal contribution. Author ordering determined by coin flip. }\quad Seungju Han\footnotemark[1]\quad Kwanghee Choi\footnotemark[1]\quad Seokjun Seo\quad Beomsu Kim\quad Buru Chang\thanks{Corresponding author.}\\
Hyperconnect\\
{\tt\small \{youngkyu.hong,seungju.han,kwanghee.choi,seokjun.seo,beomsu.kim,buru.chang\}@hpcnt.com}
}

\maketitle

\begin{abstract}\label{sec:0_abstract}
The current evaluation protocol of long-tailed visual recognition trains the classification model on the long-tailed source label distribution and evaluates its performance on the uniform target label distribution.
Such protocol has questionable practicality since the target may also be long-tailed.
Therefore, we formulate long-tailed visual recognition as a label shift problem where the target and source label distributions are different.
One of the significant hurdles in dealing with the label shift problem is the entanglement between the source label distribution and the model prediction.
In this paper, we focus on disentangling the source label distribution from the model prediction.
We first introduce a simple but overlooked baseline method that matches the target label distribution by post-processing the model prediction trained by the cross-entropy loss and the Softmax function.
Although this method surpasses state-of-the-art methods on benchmark datasets, it can be further improved by directly disentangling the source label distribution from the model prediction in the training phase.
Thus, we propose a novel method, LAbel distribution DisEntangling (LADE) loss based on the optimal bound of Donsker-Varadhan representation.
LADE achieves state-of-the-art performance on benchmark datasets such as CIFAR-100-LT, Places-LT, ImageNet-LT, and iNaturalist 2018.
Moreover, LADE outperforms existing methods on various shifted target label distributions, showing the general adaptability of our proposed method.
\end{abstract}

\section{Introduction}\label{sec:1_introduction}
\begin{figure}
\centering
\begin{center}
\includegraphics[width=\columnwidth]{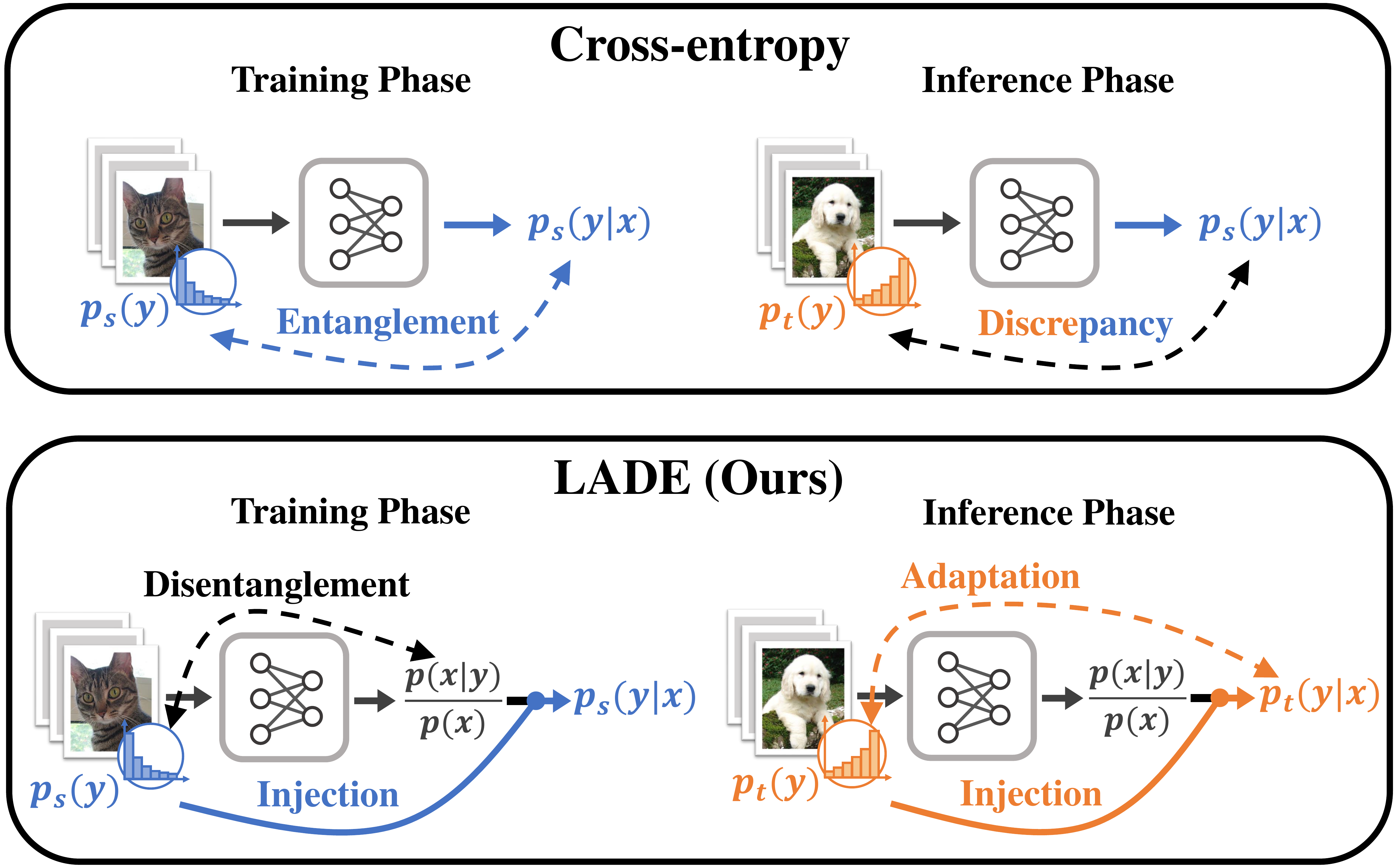}
\end{center}
\caption{
A comparison between the cross-entropy loss and our proposed LADE loss in long-tailed visual recognition.
After training with the cross-entropy loss, the model prediction gets entangled with the source label distribution $p_s(y)$, which causes a discrepancy with the target label distribution $p_t(y)$ during the inference phase.
Our proposed LADE disentangles $p_s(y)$ from the model prediction so that it can adapt to the arbitrary target probability by injecting $p_t(y)$.
}
\label{fig:1_problem_definition}
\end{figure}
Based on large-scale datasets such as ImageNet~\cite{russakovsky2015imagenet}, COCO~\cite{lin2014microsoft}, and Places~\cite{zhou2017places}, deep neural networks have achieved significant progress in various visual recognition tasks, including classification~\cite{he2016deep,simonyan2014very}, object detection~\cite{ren2015faster,girshick2015fast}, and segmentation~\cite{ronneberger2015u}.
In contrast to these relatively balanced datasets, real-world data often exhibit long-tailed distribution where head (major) classes occupy most of the data, while tail (minor) classes have a handful of samples~\cite{van2017devil,liu2019large}.
Unfortunately, the performance of state-of-the-art classification models degrades on datasets following the long-tailed distribution~\cite{buda2018systematic,he2009learning,wang2017learning}.

To tackle this problem, many long-tailed visual recognition methods~\cite{buda2018systematic,he2009learning,japkowicz2002class,byrd2019effect,shen2016relay,yang2020rethinking,cao2019learning} have been proposed.
These methods compare their effectiveness by (1) training on the long-tailed source label distribution $p_s(y)$ and (2) evaluating on the uniform target label distribution $p_t(y)$.
However, we argue that this evaluation protocol is often impractical as it is natural to assume that $p_t(y)$ could be the arbitrary distribution such as uniform distribution~\cite{russakovsky2015imagenet} and long-tailed distribution~\cite{geras2017high,carreira2018short}.

From this perspective, we are motivated to explore a new method that adapts the model to the arbitrary $p_t(y)$.
In this paper, we borrow the concept of the label distribution shift problems~\cite{garg2020unified,lipton2018detecting,tian2020posterior} to the long-tailed visual recognition task.

\begin{figure}
    \centering
    \begin{subfigure}{0.495\columnwidth}
        \centering
        \includegraphics[width=\textwidth]{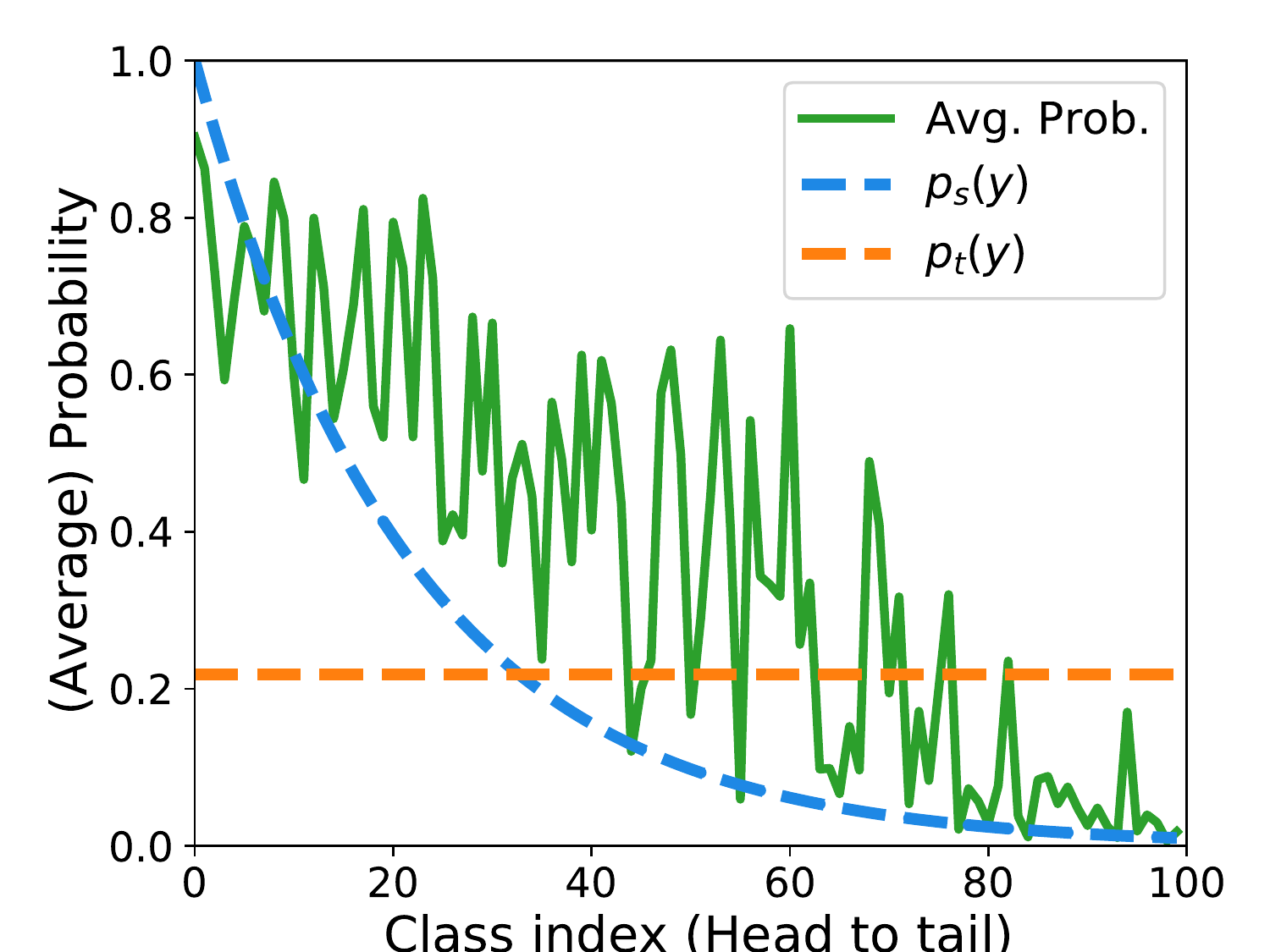}
        \caption{Cross-entropy}
        \label{fig:2_a_cross_entropy}
    \end{subfigure}
    \begin{subfigure}{0.495\columnwidth}
        \centering
        \includegraphics[width=\textwidth]{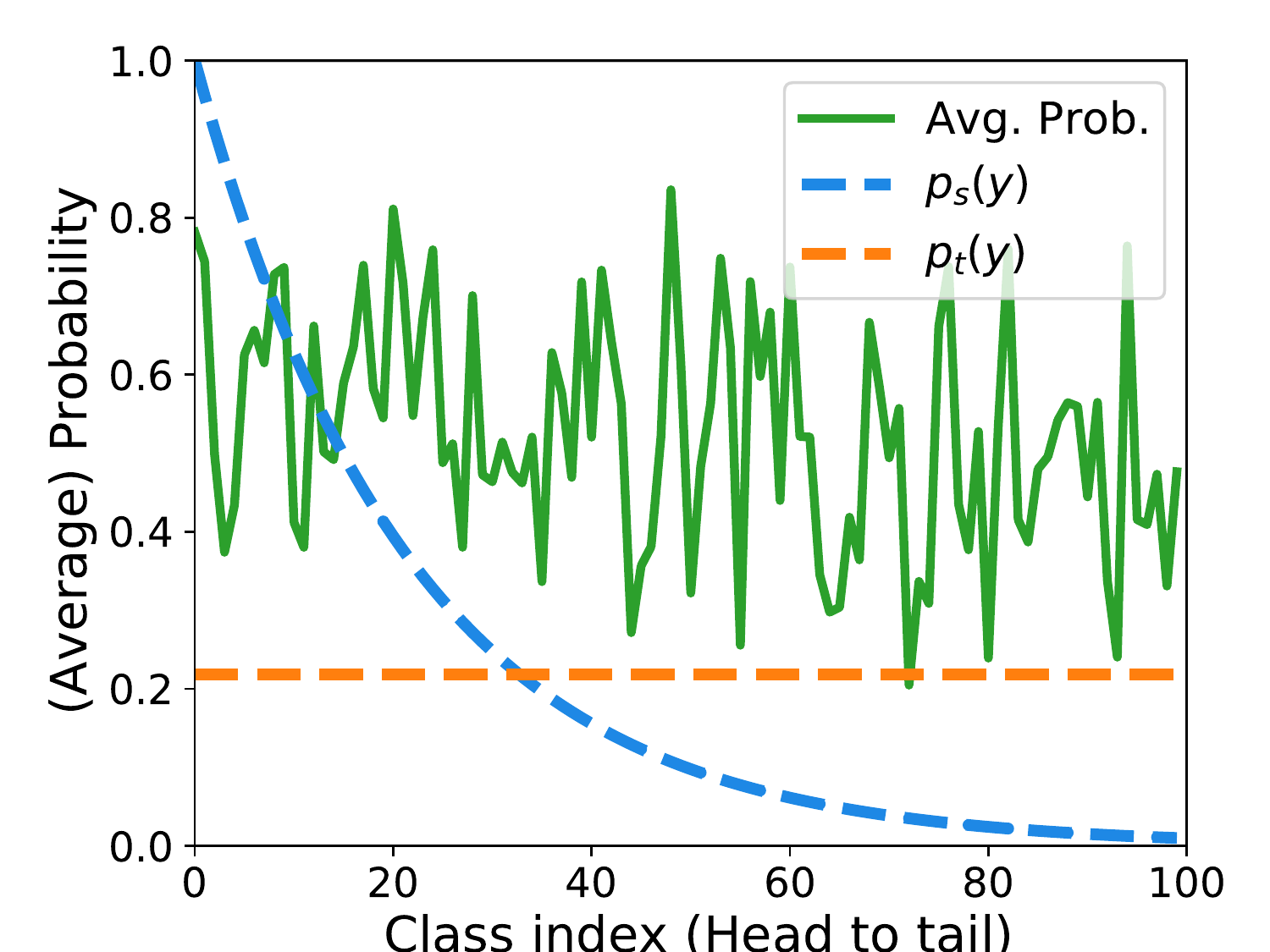}
        \caption{LADE}
        \label{fig:2_b_lade}
    \end{subfigure}
    \caption{
    The average probability for each class calculated on the balanced test set.
    The model is ResNet-32~\cite{he2016deep} trained on CIFAR-100-LT~\cite{cao2019learning}, which has uniform target label distribution $p_t(y)$ while the source dataset has long-tailed distribution $p_s(y)$.
    (a) depicts the average probability when trained with the cross-entropy loss, which shows average probability correlates with $p_s(y)$, resulting in the discrepancy with $p_t(y)$.
    (b) depicts the average probability when trained and inferred with LADE, which shows the ability of LADE on adapting to $p_t(y)$.
    }
    \label{fig:2_logit_correlation}
\end{figure}

However, it is problematic to directly use the model prediction $p(y|x;\theta)$ which is fitted to the source probability $p_s(y|x)$, as the target probability $p_t(y|x)$ is shifted from $p_s(y)$ to $p_t(y)$ (Figure~\ref{fig:1_problem_definition}).
Figure~\ref{fig:2_a_cross_entropy} shows the entanglement between the model prediction and $p_s(y)$ when the model is trained by the cross-entropy (CE) loss and the Softmax function.
To alleviate this problem, we focus on disentangling $p_s(y)$ from the model outputs so that the shifted target label distribution $p_t(y)$ can be injected to estimate the target probability.

We shed light on a simple yet strong baseline, called \textbf{P}ost-\textbf{C}ompensated Softmax (PC Softmax) that post-processes the model prediction to disentangle $p_s(y)$ from $p(y|x; \theta)$ and then incorporate $p_t(y)$ to the disentangled model output probability.
Despite the simplicity of the method, PC Softmax outperforms state-of-the-art methods in long-tailed visual recognition (will be described in Section~\ref{sec:4_experiments}).
Although this observation demonstrates the effectiveness of the disentanglement in the inference phase, PC Softmax can be further improved by directly disentangling $p_s(y)$ in the training phase.

Thus, we propose a novel method, \textbf{LA}bel distribution \textbf{D}is\textbf{E}ntangling (LADE) loss.
LADE utilizes the Donsker-Varadhan (DV) representation~\cite{donsker1985large} to directly disentangle $p_s(y)$ from $p(y|x; \theta)$.
Figure~\ref{fig:2_b_lade} shows that LADE disentangles $p_s(y)$ from $p(y|x;\theta)$.
We claim that the disentanglement in the training phase shows even better performance on adapting to arbitrary target label distributions.

We conduct several experiments to compare our proposed method with existing long-tailed visual recognition methods, and show that LADE achieves state-of-the-art performance on benchmark datasets such as CIFAR-100-LT~\cite{krizhevsky2009learning}, Places-LT~\cite{liu2019large}, ImageNet-LT~\cite{liu2019large}, and iNaturalist 2018~\cite{van2018inaturalist}.
Moreover, we demonstrate that the classification model trained with LADE can cope with arbitrary $p_t(y)$ by evaluating the performance on datasets with various shifted $p_t(y)$.
We further show that our proposed LADE can also be effective in terms of confidence calibration.
Our contributions in this paper are summarized as follows:
\begin{itemize}
    \item We introduce a simple yet strong baseline method, PC Softmax, which outperforms state-of-the-art methods in long-tailed visual recognition benchmark datasets.
    \item We propose a novel loss called LADE that directly disentangles the source label distribution in the training phase so that the model effectively adapts to arbitrary target label distributions.
    \item We show that LADE achieves state-of-the-art performance in long-tailed visual recognition on various target label distributions.
\end{itemize}
\section{Related work}\label{sec:2_related_work}
\subsection{Long-tailed visual recognition}\label{subsec:2_1_long_tailed_visual_recognition}
Most long-tailed visual recognition methods can be divided into two strategies: 
modifying the data sampler to balance the class frequency during optimization~\cite{buda2018systematic,he2009learning,japkowicz2002class,byrd2019effect,shen2016relay}, and modifying the class-wise weights of the classification loss to increase the importance of tail classes in terms of empirical risk minimization~\cite{akbani2004applying,zhou2005training,liu2006influence,margineantu2000does,japkowicz2002class,he2009learning,buda2018systematic}.

Both strategies suffer from under-representation of head classes or memorization of tail classes~\cite{yang2020rethinking,cao2019learning}. 
To overcome these problems, \cite{huang2016learning,kang2019decoupling,cao2019learning,zhou2020bbn} introduce strategies for preventing deteriorated representation learning caused by re-balancing.
\cite{wang2017learning,liu2019large} utilize knowledge from head classes to learn tail classes.
\cite{kim2020m2m,wang2020deep} augment tail class samples while preserving the diversity of dataset.
\cite{jamal2020rethinking} applies domain adaptation on learning tail class representation. 

Recent approaches introduce advanced re-balancing methods for better accommodation of tail class samples.
\cite{cui2019class} calculates the effective number of samples per each class to re-balance the loss.
\cite{cao2019learning} enforces a greater margin from the decision boundary for tail classes.
\cite{tang2020long} disentangles feature learning from the confounding effect in momentum by backdoor adjustment.

\subsection{Label distribution shift}\label{subsec:2_2_label_distribution_shift}
In this paper, we cope with long-tailed visual recognition as one of the label distribution shift problems.
We summarize recently proposed studies in label distribution shift problems that assume $p_s(y)\neq p_t(y)$ and $p_s(x|y) = p_t(x|y)$.
\cite{lipton2018detecting} estimates the degree of label shift using a black box predictor. 
\cite{garg2020unified} extends the work of ~\cite{lipton2018detecting} with the expectation-maximization algorithm.
\cite{azizzadenesheli2019regularized} introduces a domain adaptation to handle label distribution shifts by estimating importance weights.
\cite{ren2020balanced} adjusts the logits before applying the Softmax function by the frequency of each class considering the uniform target label distribution.

\subsection{Donsker-Varadhan representation}\label{subsec:2_2_Donsker_Varadhan_Loss_Function}
Donsker-Varadhan (DV) representation~\cite{donsker1985large} is the dual variational representation of Kullback-Leibler (KL) divergence~\cite{kullback1951information}. 
It is proven that the optimal bound of the DV representation is the log-likelihood ratio of two distributions of the KL divergence~\cite{banerjee2006bayesian,belghazi2018mutual}.
The usefulness of the DV representation has been broadly shown in the area including mutual information estimation~\cite{belghazi2018mutual,lin2019data,poole2019variational} or generative models~\cite{belghazi2018mutual,nowozin2016f}.
However, \cite{song2019understanding,choi2020regularized,mcallester2020formal} have pointed out the instability of directly using the DV representation.
To avoid the issue, we use the regularized DV representation from \cite{choi2020regularized} to approximate network logits as the log-likelihood ratio $\log(p(x|y)/p(x))$.
To the best of our knowledge, this is the first attempt to utilize the optimal bound inside the DV representation in the long-tailed visual recognition.
\section{Method}\label{sec:3_method}
\subsection{Preliminaries}
We start by revisiting the most common loss for training the Softmax regression (also known as the multinomial logistic regression) model~\cite{bishop2006pattern}, namely the CE loss:
\begin{align}
\label{eq:softmax}
    p(y|x;\theta) &= \frac{e^{f_\theta(x)[y]}}{\sum_{c} e^{ f_\theta(x)[c]}} \\
\label{eq:cross_entropy}
    \mathcal{L}_{CE}(f_\theta(x), y) &= -\log(p(y|x;\theta)),
\end{align}
where $x$ is the input image and $y$ is the target label, $p_s(x, y)$ and $p_t(x, y)$ are the source (train) and target (test) data distributions, and $f_\theta(x)[y]$ is the logit of class $y$ of the model. 
The Softmax regression model estimates $p_s(y|x)$ and works well \textit{when the source and the target label distribution are the same}, i.e. $p_s(y) = p_t(y)$.

However, we focus on the \textit{label shift} problem~\cite{garg2020unified,lipton2018detecting} where the target label distribution is shifted from the source label distribution, i.e. $p_s(x|y) = p_t(x|y)$ but $p_s(y) \neq p_t(y)$.
Since the model prediction estimates $p_s(y|x)$, it cannot be used to predict the shifted distribution.
This is due to the strong coupling between $p_s(y|x)$ and $p_s(y)$, as justified from the Bayes' rule:
\begin{align}
    p_s(y|x) = \frac{p_s(y)p_s(x|y)}{p_s(x)} = \frac{p_s(y)p_s(x|y)}{\sum_c p_s(c)p_s(x|c)}.
\label{eq:02_Bayes_Rule}
\end{align}

\subsection{PC Softmax: Post-Compensated Softmax}\label{subsec:3_1_baseline}
The straightforward way to handle the label distribution shift is by replacing $p_s(y)$ with $p_t(y)$.
We introduce a \textit{post-compensation} (PC) strategy that modifies the logit in the inference phase:
\begin{definition}[Post-Compensation Strategy]
\label{def:pc_strategy}
The Post-Compensation strategy modifies model logits as follows:
\begin{align}
    f_\theta^{PC}(x)[y] = f_\theta(x)[y] - \log p_s(y) + \log p_t(y) 
\end{align}
where $p_s(y)$ is the distribution which the model logits are entangled with, and $p_t(y)$ is the target distribution that the model tries to incorporate with.
\end{definition}
Note that the concept of the PC strategy is not entirely new since \cite{margineantu2000does,buda2018systematic,johnson2019survey} previously covered as a different form of multiplying $p_t(y)/p_s(y)$ to the output probability.
However, our PC strategy does not violate the categorical probability assumption, i.e. $\sum_c p_t(y=c|x)=1$.

We apply the PC strategy to the Softmax regression model, which we call Post-Compensated Softmax (PC Softmax).
For the Softmax regression model, the PC strategy is the proper adjustment for estimating target data distribution.
\begin{theorem}
\label{thm:01_post_compensated_softmax}
\normalfont{(Post-Compensated Softmax).} Let $p_s(x, y)$ and $p_t(x, y)$ be the source and target data distributions, respectively.
If $f_\theta(x)[y]$ is the logit of class $y$ from the Softmax regression model estimating $p_s(y|x)$, then the estimation of $p_t(y|x)$ is formulated as:
\begin{align}
    p_t(y|x;\theta) &= \frac {\frac{p_t(y)}{p_s(y)} \cdot e^ {f_\theta(x)[y]}}{\sum_{c} \frac{p_t(c)}{p_s(c)} \cdot e^ {f_\theta(x)[c]}} \\
    &= \frac {e^ {(f_\theta(x)[y] - \log p_s(y) + \log p_t(y))}}{\sum_{c} e^ {(f_\theta(x)[c] - \log p_s(c) + \log p_t(c))}} \\ 
    &= \frac {e^ {f_\theta^{PC}(x)[y]}}{\sum_{c} e^ {f_\theta^{PC}(x)[c]}}.
\label{eq:03_modified}
\end{align}
\textit{Proof.} See the Supplementary Material.
\end{theorem}

We emphasize that PC Softmax becomes a strong baseline that surpasses previous state-of-the-art long-tailed visual recognition methods.
However, recent literature does not consider this as a baseline.

PC Softmax can also be viewed as an extension of Balanced Softmax~\cite{ren2020balanced}, which modifies the Softmax function to accommodate the uniform target label distribution in the training phase.
In contrast, PC Softmax modifies the model logits in the inference phase to match the arbitrary target label distribution $p_t(y)$.

\subsection{LADER: LAbel distribution DisEntangling Regularizer}\label{subsec:3_2_lader}
Performance gain from the PC strategy shows the efficacy of disentangling the source label distribution.
However, the PC strategy does not involve the disentanglement in the training phase, which we claim as the ingredient for better adaptability to arbitrary target label distributions.
To achieve this, we design a new modeling objective that works as a substitute for $p_s(y|x)$. 
We derive the new objective in two steps:
(1) detaching $p_s(y)$ from $p_s(y|x)$, which results in $p_s(x|y)/p_s(x)$, and
(2) replacing $p_s(y)$ in $p_s(x)$ with the uniform prior $p_u(y)$, i.e. $p_u(y=c) = 1/C$, where $C$ is the total number of classes.

Finally, the modeling objective for the model logits is:
\begin{align}
    f_\theta(x)[y] = \log \frac {p_u(x|y)}{p_u(x)}.
\label{eq:05_logit_modeling}
\end{align}
We utilize the optimal form of the regularized Donsker-Varadhan (DV) representation~\cite{donsker1985large,banerjee2006bayesian,belghazi2018mutual,choi2020regularized} to model the log-likelihood ratio above explicitly.
\begin{theorem}
\label{thm:02_dv}
\normalfont{(Optimal form of the regularized DV representation).} Let $\mathbb{P}$, $\mathbb{Q}$ be arbitrary distributions with $supp(\mathbb{P}) \subseteq supp(\mathbb{Q})$. Suppose for every function $T: \Omega \to \mathbb{R}$ on some domain $\Omega$, the function $T$ that minimizes the regularized DV representation is the log-likelihood ratio of $\mathbb{P}$ and $\mathbb{Q}$:
\begin{equation}
    \begin{aligned}
    \log \frac{d\mathbb{P}}{d\mathbb{Q}} = \underset{T:\Omega \to \mathbb{R}}{\arg\max} (\mathbb{E}_\mathbb{P}[T] - \log(\mathbb{E}_\mathbb{Q}[e^T]) \\
    - \lambda (\log(\mathbb{E}_\mathbb{Q}[e^T]))^2),
    \end{aligned}
    \label{eq:06_DV_optimal_form}
\end{equation}
for any $\lambda \in \mathbb{R}^+$ when the expectations are finite.
\newline

\noindent \textit{Proof.} See Subsection 7.2 from~\cite{choi2020regularized}.
\end{theorem}

By plugging $\mathbb{P}=p_u(x|y)$ and $\mathbb{Q}=p_u(x)$ into Equation~\ref{eq:06_DV_optimal_form} and choosing the function family of $T: \Omega \to \mathbb{R}$ to be parametrized by the logits of the deep neural network, the optimal $f_\theta(x)[y]$ approaches to the target objective in Equation~\ref{eq:05_logit_modeling}:
\begin{equation}
    \begin{aligned}
    \log \frac {p_u(x|y)}{p_u(x)} \geq \underset{f_\theta}{\arg\max} (\mathbb{E}_{x \sim p_u(x|y)}[f_\theta(x)[y]] \\
    - \log \mathbb{E}_{x \sim p_u(x)}[e^{f_\theta(x)[y])}] \\
    - \lambda (\log (\mathbb{E}_{x \sim p_u(x)}[e^{f_\theta(x)[y])}]))^2).
    \end{aligned}
    \label{eq:07_DV_classification}
\end{equation}

Since the exact estimations of the expectation with respect to $p_u(x|y)$ and $p_u(x)$ are intractable, we use the Monte Carlo approximation~\cite{rubinstein2016simulation} using a single batch:

\begin{align}
    \mathbb{E}_{x \sim p_u(x|c)}[f_\theta(x)[c]] \approx \frac{1}{N_c} \sum_{i=1}^{N}\mathbbm{1}_{y_i=c} \cdot f_\theta(x_i)[c] \label{eq:09_montecarlo_1} \\
    \mathbb{E}_{x \sim p_u(x)}[e^{f_\theta(x)[c]}] = \mathbb{E}_{(x,y) \sim p_s(x, y)}[\frac{p_u(y)}{p_s(y)} e^ {f_\theta(x)[c]}] \label{eq:10_montecarlo_2} \\
    \approx \frac{1}{N}\sum_{i=1}^{N}\frac{p_u(y_i)}{p_s(y_i)} \cdot e^{f_\theta(x_i)[c]}  \label{eq:11_montecarlo_3},
\end{align}
where $x_i$ and $y_i$ are $i$-th sample and label, respectively, $N$ is the total number of samples, and $N_c$ is the number of samples for class $c$.
In Equation~\ref{eq:10_montecarlo_2}, importance sampling~\cite{kahn1955use} is used to approximate the expectation with respect to $p_u(x)$ using samples from $p_s(x)$:
\begin{align}
    \frac{p_u(x)}{p_s(x)} = \frac{\sum_c p_u(x|c)p_u(c)}{\sum_c p_s(x|c)p_s(c)} = \frac{p_u(y)}{p_s(y)}
     \label{eq:12_importance_sampling},
\end{align}
for the sample label pair $(x,y)\sim p_s(x,y)$, where we assume $p_s(x|c)=0$ for $c \neq y$.

Finally, we derive a novel loss that regularizes the logits to approach Equation~\ref{eq:05_logit_modeling} by applying Equation \ref{eq:09_montecarlo_1}, \ref{eq:10_montecarlo_2}, and \ref{eq:11_montecarlo_3} to Equation \ref{eq:07_DV_classification}:
\begin{definition}[LADER]
\label{def:lader}
For a single batch of sample-label pairs $(x_i, y_i)$ with $i=1, ..., N$, LAbel distribution DisEntangling Regularizer (LADER) is defined as follows:
\begin{equation}
    \begin{aligned}
    \label{eq:08_LADER_c}
    {\mathcal{L}_{LADER}}_c = & -\frac{1}{N_c}\sum_{i=1}^{N}\mathbbm{1}_{y_i=c} \cdot f_\theta(x_i)[c]\\
    &+ \log(\frac{1}{N}\sum_{i=1}^{N}\frac{p_u(y_i)}{p_s(y_i)} \cdot e^{f_\theta(x_i)[c]})\\
    &+ \lambda (\log(\frac{1}{N}\sum_{i=1}^{N} \frac{p_u(y_i)}{p_s(y_i)} \cdot e^ {f_\theta(x_i)[c]})) ^ 2\\
    \end{aligned}
\end{equation}
\begin{equation}
    \label{eq:09_LADER}
    \mathcal{L}_{LADER} = \sum_{c \in \mathbb{S}}{\alpha_c \cdot {\mathcal{L}_{LADER}}_c},
\end{equation}
with nonnegative hyperparameters $\lambda$, $\alpha_1$, $\dots$, $\alpha_C$, where $C$ is total number of classes, $N_c$ is the number of samples of class $c$ and $\mathbb{S}$ is the set of classes existing inside the batch.
\end{definition}
Empirically, we find out that regularizing the head classes more strongly than the tail classes is more effective.
Thus, we apply $\alpha_c=p_s(y=c)$ as the weight for the regularizer of class $c$, ${\mathcal{L}_{LADER}}_c$ in Equation~\ref{eq:09_LADER}.

\subsection{Deriving the conditional probability from disentangled logits}\label{subsec:3_3_cond_prob}
LADER regularizes the logits to be $\log (p_u(x|y)/p_u(x))$ to ensure the logits are explicitly disentangled from the source label distribution $p_s(y)$.
To estimate the conditional probability $p_t(y|x)$ of the arbitrary data distribution $p_t(x, y)$ from the regularized logits, we use the modified Softmax function derived from the Bayes’ rule with the assumption of $p_t(x|y) = p_u(x|y)$:
\begin{equation}
\begin{aligned}
\label{eq:cond_prob_lader}
    &p_t(y|x;\theta)
    = \frac{p_t(y)p_t(x|y;\theta)}{\sum_c p_t(c)p_t(x|c;\theta)}\\
    &= \frac{p_t(y)p_u(x|y;\theta)}{\sum_c p_t(c)p_u(x|c;\theta)} 
= \frac{p_t(y) \cdot e^{f_\theta(x)[y]}}{\sum_c p_t(c) \cdot e^{f_\theta(x)[c]}}.
\end{aligned}
\end{equation}

Similar to this, we can estimate $p_s(y|x)$ by swapping $p_t(y)$ of Equation~\ref{eq:cond_prob_lader} with $p_s(y)$, so that $p_s(y|x;\theta)$ can be optimized by the CE loss.
Thus, we can combine LADER with the CE loss as our final loss for training:
\begin{definition}[LADE]
\label{def:lade}
LAbel distribution DisEntangling (LADE) loss is defined as follows:
\begin{align}
    \mathcal{L}_{LADE-CE}(f_\theta(x), y) = -\log(p_s{(y|x;\theta)}) \\
    = -\log(\frac{p_s(y) \cdot e^{f_\theta(x)[y]}}{\sum_c p_s(c) \cdot e^{f_\theta(x)[c]}}) \label{eq:ce_lader}
\end{align}
\begin{equation}
\begin{aligned}
    \mathcal{L}_{LADE}(f_\theta(x), y) =   \mathcal{L}_{LADE-CE}(f_\theta(x), y) \\ + \alpha \cdot \mathcal{L}_{LADER}(f_\theta(x), y) \label{eq:final_loss},
\end{aligned}
\end{equation}
where $\alpha$ is a nonnegative hyperparameter, which determines the regularization strength of $\mathcal{L}_{LADER}$.
\end{definition}

Note that Balanced Softmax~\cite{ren2020balanced} is equivalent to LADE with $\alpha=0$, but LADE is derived from an entirely different perspective of directly regularizing the logits.
Furthermore, Balanced Softmax only covers the uniform target label distribution, while our method is designed to cover arbitrary target label distributions without re-training.

In the inference phase, we inject the target label distribution as in Equation~\ref{eq:cond_prob_lader}.
\section{Experiments}\label{sec:4_experiments}
We compare PC Softmax and LADE with current state-of-the-art methods.
First, we evaluate the performance on the uniform target label distribution, which is the prevalent evaluation scheme of long-tailed visual recognition. 
Then, we assess the performance on variously shifted target label distributions.
Finally, we conduct further analysis to show that LADE successfully disentangles the source label distribution and improves confidence calibration.
We provide source codes\footnote{\url{https://github.com/hyperconnect/LADE}} of LADE for the reproduction of the experiments conducted in this paper.
Details of the hyperparameter tuning process and the results of the ablation test are reported in the Supplementary Material.

\subsection{Experimental setup}\label{sec:4_1_experimental_setup}
\paragraph{Long-tailed dataset}
We follow the common evaluation protocol~\cite{liu2019large, cao2019learning,zhou2020bbn,tang2020long} in long-tailed visual recognition, which trains classification models on the long-tailed source label distribution and evaluates their performance on the uniform target label distribution.
We use four benchmark datasets with at least 100 classes to simulate the real-world long-tailed data distribution: CIFAR-100-LT~\cite{cao2019learning}, Places-LT~\cite{liu2019large}, ImageNet-LT~\cite{liu2019large}, and iNaturalist 2018 \cite{van2018inaturalist}.
We define the \textit{imbalance ratio} as $N_{max} / N_{min}$, where $N$ is the number of samples in each class.
CIFAR-100-LT has three variants with controllable data imbalance ratios $10$, $50$, and $100$.
The details of datasets are summarized in Table~\ref{tab:dataset}.

\paragraph{Comparison with other methods.} 
We compare PC Softmax and LADE with three categories of methods:
\begin{itemize}[leftmargin=*]
\item \textbf{Baseline methods}. For our baseline, we use Softmax (Equation~\ref{eq:softmax}), Focal loss (Focal)~\cite{lin2017focal}, OLTR~\cite{liu2019large}, CB-Focal~\cite{cui2019class}, and LDAM~\cite{cao2019learning}.

\item \textbf{Two-stage training}.
To demonstrate our method's efficiency and effectiveness, we compare our method with two-staged state-of-the-art methods that employ a fine-tuning strategy.
LDAM+DRW~\cite{cao2019learning} applies a fine-tuning step with loss re-weighting.
Decouple~\cite{kang2019decoupling} re-balances the classifier during the fine-tuning stage.

\item \textbf{Other state-of-the-art methods}. 
BBN~\cite{zhou2020bbn}, Causal Norm~\cite{tang2020long}, and Balanced Softmax~\cite{ren2020balanced} are recently proposed state-of-the-art methods on long-tail visual recognition.
BBN uses an extra additional network branch to deal with an imbalanced training set.
Causal Norm utilizes backdoor adjustment to remove indirect causal effect caused by imbalanced source label distribution.
\end{itemize}

\begin{table}[t]
\footnotesize
\centering
\caption{The details of the training set of long-tailed datasets.}
\vspace{0pt}
{
\begin{tabular}{l c c c}
\toprule
Dataset & \# of classes & \# of samples & Imbalance ratio \\
\midrule
CIFAR-100-LT & 100 & 50K & \{10, 50, 100\} \\
Places-LT & 365 & 62.5K & 996 \\
ImageNet-LT & 1K & 186K & 256 \\
iNaturalist 2018 & 8K & 437K & 500 \\
\bottomrule
\end{tabular}
}
\label{tab:dataset}
\vspace{0mm}
\end{table}
\paragraph{Evaluation Protocol.}
We report evaluation results using top-1 accuracy.
Following~\cite{liu2019large}, for ImageNet-LT and Places-LT, we categorize the classes into three groups depending on the number of samples of each class and further report each group's evaluation results.
The three groups are defined as follows:
\textit{Many} covering classes with $> 100$ images, \textit{Medium} covering classes with $\geq 20$ and $\leq 100$ images, \textit{Few} covering classes with $< 20$ images.

\subsection{Results on balanced test label distribution}
\label{subsec:4_2_results_on_balanced_test_label_distribution}
Evaluation results on CIFAR-100-LT, Places-LT, ImageNet-LT, and iNaturalist 2018 are shown in Table~\ref{tab:cifar_main}, \ref{tab:place_main}, \ref{tab:imagenet_main}, and \ref{tab:inat_main}, respectively.
All the datasets have a uniform target label distribution.
PC Softmax shows comparable or better results than the previous state-of-the-art results on benchmark datasets.
This result is quite surprising considering the simplicity of PC Softmax.
Our proposed method, LADE, achieves even better performance in long-tailed visual recognition on all four benchmark datasets, advancing the state-of-the-art even further.

\paragraph{CIFAR-100-LT}
\begin{table}[t]
\footnotesize
\centering
\caption{Top-1 accuracy on CIFAR-100-LT with different imbalance ratios. Rows with \(\dagger\)~denote results directly borrowed from~\cite{tang2020long}. We use the same backbone network with~\cite{tang2020long}.}
\vspace{0pt}
{
\begin{tabular}{l  c  c  c}
\toprule
Dataset & \multicolumn{3}{c}{CIFAR-100 LT} \\ 
\midrule
Imbalance ratio & 100 & 50 & 10 \\
\midrule
Focal Loss\(^\dagger\) & 38.4 & 44.3 & 55.8 \\
LDAM\(^\dagger\) & 42.0 & 46.6 & 58.7 \\
BBN\(^\dagger\) & 42.6 & 47.0 & 59.1 \\
Causal Norm\(^\dagger\) & 44.1 & 50.3 & 59.6 \\
Balanced Softmax & 45.1 & 49.9 & 61.6 \\
Softmax & 41.0 & 45.5 & 59.0 \\
\midrule
PC Softmax & 45.3 & 49.5 & 61.2 \\
\textbf{LADE} & \textbf{45.4} & \textbf{50.5} & \textbf{61.7} \\
\bottomrule
\end{tabular}
}
\label{tab:cifar_main}
\vspace{0mm}
\end{table}
Table~\ref{tab:cifar_main} shows the evaluation results on CIFAR-100-LT.
As shown in the table, in CIFAR-100-LT, LADE outperforms all the baselines over all the imbalance ratios.
PC Softmax also shows better performance than other methods except for Balanced Softmax and our proposed LADE.

\paragraph{Places-LT}
\begin{table}
\footnotesize
\centering
\caption{The performances on Places-LT~\cite{liu2019large}, starting from an ImageNet pre-trained ResNet-152. Rows with \(\dagger\)~denote results directly borrowed from~\cite{kang2019decoupling}.}
\label{tab:place_main}
\vspace{0pt}
\begin{tabular}{l|ccc|c}
\toprule
Method & Many & Medium & Few & \textbf{All} \\
\midrule
Focal Loss\(^\dagger\) & 41.1 & 34.8 & 22.4 & 34.6 \\
OLTR\(^\dagger\) & 44.7 & 37.0 & 25.3 & 35.9 \\
Decouple-$\tau$-norm\(^\dagger\) & 37.8 & \textbf{40.7} & 31.8 & 37.9 \\
Decouple-LWS\(^\dagger\) & 40.6 &	39.1 &	28.6 & 37.6 \\
Causal Norm & 23.8 & 35.8 & \textbf{40.4} & 32.4 \\
Balanced Softmax & 42.0	& 39.3	& 30.5 & 38.6 \\
Softmax & \textbf{46.4} & 27.9 & 12.5 & 31.5 \\\midrule
PC Softmax & 43.0 & 39.1 & 29.6 & 38.7 \\
\textbf{LADE} & 42.8 & 39.0 & 31.2 & \textbf{38.8} \\
\bottomrule
\end{tabular}
\end{table}
We further evaluate PC Softmax and LADE on Places-LT, and Table~\ref{tab:place_main} shows the experimental results.
LADE achieves a new state-of-the-art of 38.8\% top-1 overall accuracy, without using a two-stage training as in Decouple-$\tau$-norm~\cite{kang2019decoupling}.
PC Softmax shows yet another promising result by surpassing the previous state-of-the-art, while Softmax offers poor results.
This result is quite impressive since both models are the same, and the only difference occurs in the inference phase.

\paragraph{ImageNet-LT}
\label{subsec:imagenet_lt}
\begin{table}[t]
\footnotesize
\centering
\caption{The performances on ImageNet-LT~\cite{liu2019large}. Rows with \(\mathsection\) denote results directly borrowed from~\cite{tang2020long}.}
\vspace{0px}
{
\begin{tabular}{l |c c c| c }
\toprule
Method & Many & Medium & Few & \textbf{All} \\ 
\midrule
\textit{90 epochs} & & & & \\
Focal Loss\(^\mathsection\) & 64.3 & 37.1 & 8.2 & 43.7 \\
OLTR\(^\mathsection\) & 51.0 & 40.8 & 20.8 & 41.9       \\
Decouple-cRT\(^\mathsection\) & 61.8 & 46.2 & 27.4 & 49.6 \\
Decouple-\(\tau\)-norm\(^\mathsection\) & 59.1 & 46.9 & 30.7 & 49.4\\
Decouple-LWS\(^\mathsection\) & 60.2 & 47.2 & 30.3 & 49.9\\
Causal Norm\(^\mathsection\) & 62.7 & 48.8 & 31.6 & 51.8 \\
Balanced Softmax & 62.2	& 48.8 & 29.8 & 51.4 \\
Softmax & 65.1 & 35.7 & 6.6 & 43.1 \\\midrule
PC Softmax & 60.4 & 46.7 & 23.8 & 48.9 \\
\textbf{LADE} & 62.3 & \textbf{49.3} & 31.2 & 51.9 \\
\midrule
\midrule
\textit{180 epochs} & & & & \\
Causal Norm & 65.2 & 47.7 & 29.8 & 52.0 \\
Balanced Softmax & 63.6 & 48.4 & 32.9 & 52.1 \\
Softmax & \textbf{68.1} & 41.9 & 14.4 & 48.2 \\\midrule
PC Softmax & 63.9  & 49.1 & \textbf{34.3} & 52.8 \\
\textbf{LADE} & 65.1 & 48.9 & 33.4 & \textbf{53.0} \\
\bottomrule
\end{tabular}
}
\label{tab:imagenet_main}
\vspace{0mm}
\end{table}
We conduct experiments on ImageNet-LT to demonstrate the effectiveness of LADE in the large-scale dataset.
We observe that the model is under-fitting at 90 epochs when using LADE.
Previous works~\cite{kang2019decoupling, zhou2020bbn} train model for longer epochs to deal with under-fitting.
Thus, we also report the evaluation results at both 90 and 180 epochs, respectively, and this is different from \cite{tang2020long} where they train the baseline methods during 90 epochs.
Table~\ref{tab:imagenet_main} presents the performance of our method on ImageNet-LT.
LADE yields 53.0\% top-1 overall accuracy with 180 epochs, which is better than the previous state-of-the-art, including Causal Norm trained with 180 epochs.
PC Softmax also shows a favorable result, 52.8\%, where it also outperforms the previous state-of-the-art-results.
Besides, LADE still achieves the best result compared to the methods when LADE and the methods are trained for 90 epochs.

\paragraph{iNaturalist 2018}
\begin{table}
\footnotesize
\centering
\caption{Top-1 accuracy over all classes on iNaturalist 2018. Rows with \(\dagger\)~denote results directly borrowed from~\cite{kang2019decoupling} and $^\star$~denotes the result directly borrowed from~\cite{zhou2020bbn}.}
\label{tab:inat_main}
\vspace{0pt}
\begin{tabular}{lc}
\toprule
Method & Top-1 Accuracy \\
\midrule
CB-Focal\(^\dagger\) & 61.1 \\
LDAM\(^\dagger\)  & 64.6 \\
LDAM+DRW\(^\dagger\) & 68.0 \\
Decouple-$\tau$-norm\(^\dagger\) & 69.3 \\
Decouple-LWS\(^\dagger\) & 69.5 \\
BBN$^\star$ & 69.6 \\
Causal Norm & 63.9 \\
Balanced Softmax & 69.8 \\
Softmax  & 65.0 \\
\midrule
PC Softmax & 69.3 \\
\textbf{LADE } & \textbf{70.0} \\
\bottomrule
\end{tabular}
\end{table}
\begin{table*}[t]
\footnotesize
\centering
\caption{Top-1 accuracy over all classes on test time shifted ImageNet-LT. All models are trained for 180 epochs.}
\vspace{0px}
{
	\begin{tabular}{l|ccccc|c|ccccc}
    \toprule
    Dataset & \multicolumn{5}{c|}{Forward} & Uniform & \multicolumn{5}{c}{Backward}\\
    \midrule
    Imbalance ratio & 50 & 25 & 10 & 5 & 2 & 1 & 2 & 5 & 10 & 25 & 50 \\
    \midrule
	Causal Norm & 64.1 & 62.5 & 60.1 & 57.8 & 54.6 & 52.0 & 49.3 & 45.8 & 43.4 & 40.4 & 38.4 \\
	Balanced Softmax & 62.5 & 60.9 & 58.8 & 57.0 & 54.4 & 52.1 & 49.6 & 46.5 & 44.1 & 41.4 & 39.7 \\
	Softmax & 66.3 & 63.9 & 60.4 & 57.1 & 52.3 & 48.2 & 44.2 & 38.9 & 35.0 & 30.5 & 27.9 \\
    \midrule
	PC Causal Norm & 66.7 & 64.3 & 60.9 & 58.1 & 54.6 & 52.0 & 49.8 & 47.9 & 47.0 & 46.7 & 46.7 \\
	PC Balanced Softmax & 65.5 & 63.1 & 59.9 & 57.3 & 54.3 & 52.1 & 50.2 & 48.8 & 48.3 & 48.5 & 49.0 \\
	PC Softmax & 66.6 & 63.9 & 60.6 & 58.1 & 55.0 & 52.8 & 51.0 & 49.3 & 48.8 & 48.5 & 49.0 \\
    \midrule
	\textbf{LADE} & \textbf{67.4} & \textbf{64.8} & \textbf{61.3} & \textbf{58.6} & \textbf{55.2} & \textbf{53.0} & \textbf{51.2} & \textbf{49.8} & \textbf{49.2} & \textbf{49.3} & \textbf{50.0} \\
	\bottomrule
	\end{tabular}
}
\label{tab:imagenet_test_shift}
\vspace{0mm}
\end{table*}
To show the scalability of LADE on a large-scale dataset, we evaluate our methods in the real-world long-tailed dataset, iNaturalist 2018.
Since iNaturalist 2018 does not contain a validation set, we train the model with LADE for 200 epochs and report the test accuracy at 200 epochs, following~\cite{kang2019decoupling}.
Table~\ref{tab:inat_main} shows the top-1 accuracy over all classes on iNaturalist 2018.
LADE reaches the best accuracy 70.0\% among the other methods, even without any branch structure of fine-tuning as BBN~\cite{zhou2020bbn}, nor a two-stage training scheme as~\cite{kang2019decoupling}.
Further, LADE surpasses PC Softmax with a large gap, +0.7\%, where PC Softmax still shows a competitive result compared to other methods.
This result indicates that PC Softmax is effective for small datasets but performance worsens for larger datasets, while LADE scales well on large datasets as well.

\subsection{Results on variously shifted test label distributions}\label{subsec:4_3_results_on_imbalanced_test_label_distribution}
Test sets are rarely well-balanced in real-world scenarios.
To simulate and compare the performance of various state-of-the-art methods in the wild, we propose a more realistic evaluation protocol. 
We first train the model on the long-tailed source label distribution.
Then we examine the performance on a range of target label distributions, from distributions that resemble the source label distributions to radically different distributions, similar to~\cite{azizzadenesheli2019regularized, lipton2018detecting}.

We choose the ImageNet-LT from Section~\ref{subsec:imagenet_lt} as the source dataset. 
The test set of ImageNet-LT is uniformly distributed, and each class has 50 samples; hence the maximum imbalance ratio is 50.
Similar to constructing the CIFAR-LT training dataset~\cite{cui2019class, cao2019learning}, we additionally design two types of test datasets. 
Let us assume ImageNet-LT classes are sorted by descending values of the number of samples per class.
Then the shifted test dataset is defined as follows:
(1) \textbf{Forward}. $n_j = N\cdot\mu^{(j - 1) / C}$. As the imbalance ratio increases, it becomes similar to the source label distribution.
(2) \textbf{Backward}. $n_j = N\cdot\mu^{(C -j) / C}$. The order is flipped so that it gets more different as the imbalance ratio increases.
Here $\mu$ is the imbalance ratio, $N$ is the number of samples per class in the original ImageNet-LT test set ($=50$), $C$ is the number of classes, $j$ is the class index and $1 \leq j \leq C$, and $n_j$ is the number of samples in class $j$ for the shifted test set.

We compare LADE with Softmax, Balanced Softmax, and Causal Norm on this evaluation protocol.
For a fair comparison, we apply our PC strategy to state-of-the-art methods: $\log p_t(y) - \log p_u(y)$ is added to the logits before applying the softmax function for Balanced Softmax, and Causal Norm, as both methods target the uninformative prior. 
Theorem~\ref{thm:01_post_compensated_softmax} is used for the Softmax instead.

Table~\ref{tab:imagenet_test_shift} shows the top-1 accuracy in the range of test datasets between the Forward- and Backward-type datasets.
Our PC strategy shows consistent performance gain, which indicates the benefits of plug-and-play target label distributions.
Moreover, LADE outperforms all the other methods in every imbalance settings, and the performance gap between LADE and PC Softmax gets wider as the dataset gets more imbalanced.
These results demonstrate the general adaptability of our proposed method on real-world scenarios, where the target label distribution is variously shifted.

\subsection{Further analysis}\label{subsec:4_4_further_analysis}
\paragraph{Visualization of the logit values}
\begin{figure*}
\begin{center}
\includegraphics[width=\textwidth]{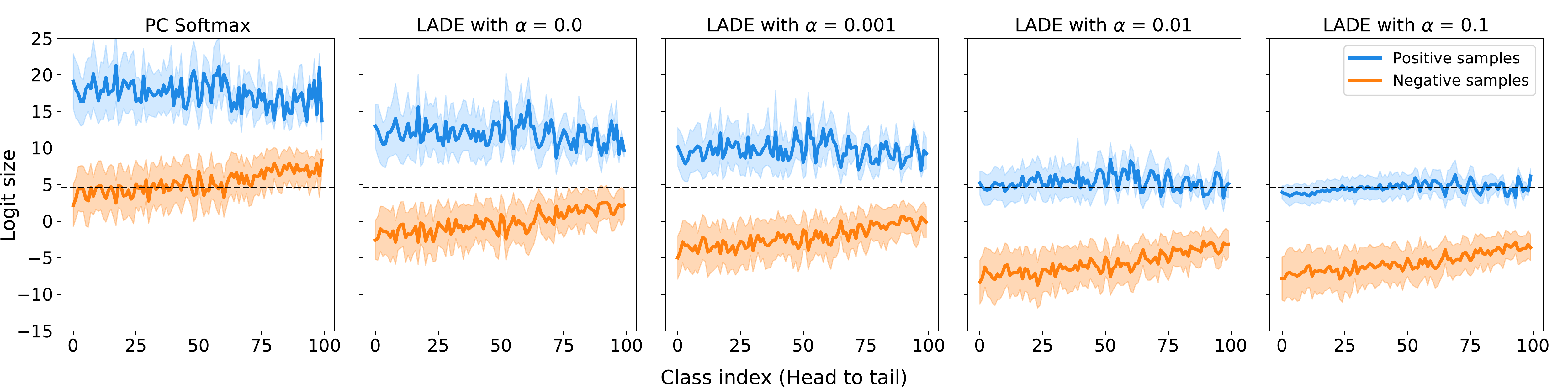}
\end{center}
\caption{
ResNet-32 model logits correspond to each class, where the model is trained on CIFAR-100-LT with an imbalance ratio of 100.
For each class $c$, \textit{positive samples} denote the sample corresponds to class $c$, and \textit{negative samples} denote the other.
The colored area denotes the variance of logit values, while the line indicates the mean.}
\label{fig:3_alpha_search}
\end{figure*}
In this subsection, we visualize the logit values of each class in order to demonstrate the effect of LADE.
By disentangling the source label distribution with LADE as described in Section~\ref{subsec:3_2_lader}, the logit value $f_\theta(x)[y]$ should converge to $\log C$ for the positive samples:
\begin{align}
    f_\theta(x)[y] = \log \frac{p_u(x|y)}{p_u(x)} = \log \frac{p_u(y|x)}{p_u(y)} = \log C,
    \label{eq:prior_disentangling}
\end{align}
where we assume perfectly separable case, i.e. $p_u(y|x) = 1$ for $(x, y) \sim p_u(x, y)$ and $C$ is the number of classes.

Figure~\ref{fig:3_alpha_search} shows how the logits are distributed for each class.
The hyperparameter $\alpha$ represents the regularization strength for LADER.
As $\alpha$ increases, the logit values gradually converge to the theoretical value $y = \log C = \log 100$ (dotted line in the figure), reconfirming Theorem~\ref{thm:02_dv}.
This result indicates that LADER successfully regularizes the logit values as we intended.

\paragraph{Confidence calibration}
\begin{figure*}
\begin{center}
\includegraphics[width=\textwidth]{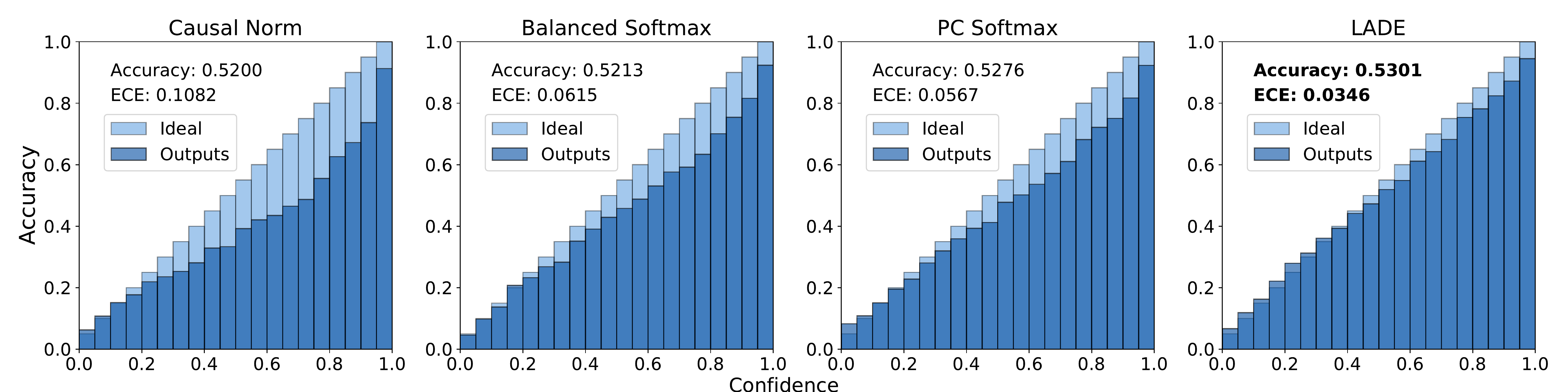}
\end{center}
\caption{Reliability diagrams of ResNeXt-50-32x4d~\cite{xie2017aggregated} on ImageNet-LT.
The average confidence of the model trained by LADE nearly matches its accuracy.
}
\label{fig:4_confidence_calibration}
\end{figure*}
Previous literature claims that the confidence of the neural network classifier does not represent its true accuracy~\cite{guo2017calibration}.
We regard a classifier is well-calibrated when its predictive probability, $\max_y p(y|x)$, represents the true probability~\cite{guo2017calibration}.
For example, when a calibrated classifier predicts the label $y$ with predictive probability 0.75, it has a 75\% chance of being correct.
It will be catastrophic if we cannot trust the confidence of the neural network in the domain of medical diagnosis~\cite{caruana2015intelligible} and self-driving car~\cite{bojarski2016end}.

\cite{muller2019does} suspect the endlessly growing logit values induced by a combination of naive CE loss and the Softmax function as the culprit of over-confidence.
Since LADER regularizes the logit size, we expect that using LADE prevents the model from being over-confident.
Through experiments, we observe that LADE improves calibration.
Figure~\ref{fig:4_confidence_calibration} shows the reliability diagrams with 20-bins.
Using expected calibration error (ECE)~\cite{naeini2015obtaining}, we quantitatively measure the miscalibration rate of the model trained on ImageNet-LT.
We compare our LADE with PC Softmax and the current state-of-the-art methods, Causal Norm~\cite{tang2020long}, and Balanced Softmax~\cite{ren2020balanced}.
Results show that LADE produces a more calibrated classifier than other methods, with the ECE of $0.0346$, which confirms our expectation.
\section{Conclusion}\label{sec:5_conclusion}
In this paper, we suggest that disentangling the source label distribution from the model prediction is useful for long-tailed visual recognition.
To disentangle the source label distribution, we first introduce a simple yet strong baseline, called PC Softmax, that matches the target label distribution by post-processing the model prediction trained by the cross-entropy loss and the Softmax function.
We further propose a novel loss, LADE, that directly disentangles the source label distribution in the training phase based on the optimal bound of Donsker-Varadhan representation.
Experiment results demonstrate that PC Softmax and our proposed LADE outperform state-of-the-art long-tailed visual recognition methods on real-world benchmark datasets.
Furthermore, LADE achieves state-of-the-art performance on various shifted target label distributions.
Lastly, further experiments show that our proposed LADE is also effective in terms of confidence calibration.
We plan to extend our research to other vision domain problems that suffer from long-tailed distributions, such as object detection and segmentation.
\clearpage
{\small
\bibliographystyle{ieee_fullname}
\bibliography{egbib}
}

\clearpage
\section*{Appendix}\label{sec:6_appendix}
\setcounter{section}{6}
\setcounter{equation}{21}
\setcounter{figure}{5}
\setcounter{table}{6}

\subsection{Proof to Theorem 1}
Assume $p_t(y|x)$ to be the target conditional probability and $p_s(y|x)$ to be the source conditional probability. We start with $p_s(y|x)$ formulated with logits $f_\theta(x)[y]$:
\begin{align}
    p_s(y|x) = \frac{e^ {f_\theta(x)[y]}}{\sum_c e^ {f_\theta(x)[c]}}.
\end{align}
By applying the $\log$ function on both sides,
\begin{equation}
\begin{aligned}
    f_\theta(x)[y] &= \log p_s(y|x) + C_x \\ 
    &= \log \left(\frac{p_s(y)p_s(x|y)}{\sum_c p_s(c)p_s(x|c)}\right) + C_x \\
    &= \log (p_s(y)p_s(x|y)) + C'_x \\
    &= \log (p_s(y)p_t(x|y)) + C'_x \\
    &= \log (p_t(y)p_t(x|y)) + \log p_s(y) \\
    &\quad-\log p_t(y) + C'_x,
\end{aligned}
\end{equation}
where $C_x$ and $C'_x$ can be regarded as constants for a fixed $x$ as follows:
\begin{align}
    C_x &= \log \left(\sum_c e^{f_\theta(x)[c]}\right), \\
    C'_x &= C_x - \log \left(\sum_c p_s(c) p_s(x|c)\right).
\end{align}
Let us derive the post-compensated logit $f_\theta^{PC}$ (Definition 3.1) from $f_\theta$:
\begin{equation}
\begin{aligned}
    \log &(p_t(y)p_t(x|y))  \\
    &= f_\theta(x)[y] - \log p_s(y) + \log p_t(y) - C'_x \\
    &= f_\theta^{PC}(x)[y] - C'_x.
\end{aligned}
\end{equation}
Re-calculating the Softmax function yields:
\begin{equation}
\begin{aligned}
    \frac{
        e^ {f_\theta^{PC}(x)[y]}
    }{
        \sum_c e^{ f_\theta^{PC}(x)[c]}
    } &= \frac{
        e^ {f_\theta^{PC}(x)[y] - C'_x}
    }{
        \sum_c e^{ f_\theta^{PC}(x)[c] - C'_x}
    } \\
    &= \frac{p_t(y)p_t(x|y)}{
        \sum_c p_t(c)p_t(x|c)
    } \\ 
    &= p_t(y|x),
\end{aligned}
\end{equation}
which ends the proof.

\subsection{Implementation details}
For all the experiments over multiple datasets, we use the SGD optimizer with momentum $\gamma = 0.9$ and weight decay $5 \cdot 10^{-4}$ to optimize the network if not specified.
We use the same random seed throughout the whole experiment for a fair comparison.
For image classification on CIFAR-100-LT and ImageNet-LT, we follow most of the details from \cite{tang2020long}, and on Places-LT and iNaturalist 2018, we follow~\cite{kang2019decoupling}.
All the models are trained on 4 GPUs, except CIFAR-100-LT, where we use 1 GPU.
We find the optimal hyperparameters based on a grid search with the validation set. 
However, as the iNaturalist 2018 dataset does not contain the validation set, we use the same $\lambda$ and $\alpha$ searched on the ImageNet-LT dataset since it has a similar number of classes and samples compared to the iNaturalist 2018 dataset.
Detailed experiment settings for LADE are summarized in Table~\ref{tab:07_hparams}.
\begin{table}[h]
\footnotesize
\centering
\caption{Experimental settings on four benchmark datasets when using LADE. \textit{IB} stands for the imbalance ratio.}
\vspace{0pt}
{
\begin{tabular}{l c c c}
\toprule
Dataset & $\lambda$ & $\alpha$ & Batch size \\
\midrule
CIFAR-100-LT (\textit{IB} 10) & 0.01 & 0.01 & 256 \\
CIFAR-100-LT (\textit{IB} 50) & 0.01 & 0.01 & 256 \\
CIFAR-100-LT (\textit{IB} 100) & 0.01 & 0.1 & 256 \\
Places-LT & 0.1 & 0.005 & 128\\
ImageNet-LT & 0.5 & 0.05 & 256 \\
iNaturalist 2018 & 0.5 & 0.05 & 256 \\
\bottomrule
\end{tabular}
}
\label{tab:07_hparams}
\vspace{0mm}
\end{table}

\paragraph{CIFAR-100-LT \cite{krizhevsky2009learning}}
On the CIFAR-100-LT dataset, we use ResNet-32~\cite{he2016deep} as the backbone network for all the experiments, following the implementation of \cite{tang2020long}.
We train for 200 epochs and apply the linear warm-up learning rate schedule~\cite{goyal2017accurate} to the first five epochs.
The learning rate is initialized as $0.2$, and it is decayed at the 120th and 160th epoch by $0.01$.

\paragraph{Places-LT \cite{zhou2017places}}
We use ResNet-152~\cite{he2016deep} as the backbone network with pretraining on the ImageNet-2012~\cite{deng2009imagenet} dataset.
We use 0.05 and 0.001 for the initial learning rate of the classifier and the feature extractor.
We train for 30 epochs with a learning rate decay of $0.1$ every 10 epochs.

\paragraph{ImageNet-LT \cite{deng2009imagenet}}
On the ImageNet-LT dataset, we utilize ResNeXt-50-32x4d~\cite{xie2017aggregated} as the backbone network for all the experiments.
We use the cosine learning rate schedule~\cite{loshchilov2016sgdr} decaying from $0.05$ to $0.0$ during 180 epochs.

\paragraph{iNaturalist 2018 \cite{van2018inaturalist}}
For the iNaturalist 2018 dataset, we use ResNet-50~\cite{he2016deep} as the backbone network for all experiments.
We use cosine learning rate scheduling~\cite{loshchilov2016sgdr} decaying from $0.1$ to $0.0$ during 200 epochs, following~\cite{kang2019decoupling}.

\paragraph{Data Pre-processing}
We follow~\cite{liu2019large} for the details on image preprocessing.
For the training set, images are resized to 256 $\times$ 256 and randomly cropped to 224 $\times$ 224.
After cropping, we augment images with random horizontal flip with probability $p = 0.5$ and apply random color jitter.
For validation and test set, images are center cropped to 224 $\times$ 224 without any augmentation.

\subsection{Ablation study}
To verify the effectiveness of the regularizer term for DV representation (Equation 9) and LADER (Equation 16), we conduct an ablation test.
Table~\ref{tab:ablation} shows how the top-1 accuracy changes when removing the regularizer term for the DV representation ($\lambda = 0$) or removing LADER ($\alpha = 0$), respectively.
\begin{table}[h]
\footnotesize
\centering
\caption{Ablation study for LADE on the long-tailed benchmark datasets.
LADE (Ours) shows the best evaluation performance, and $\lambda = 0$ and $\alpha = 0$ denote the performance with the same settings except for the DV representation regularization or LADER, respectively.}

\vspace{0pt}
{
\begin{tabular}{l c c c}
\toprule
Dataset & LADE (Ours) & $\lambda = 0$ & $\alpha = 0$ \\
\midrule
CIFAR-100-LT (\textit{IB} 10) & \textbf{61.7} & 61.5 & 61.6\\
CIFAR-100-LT (\textit{IB} 50) & \textbf{50.5} & 49.5 & 49.9\\
CIFAR-100-LT (\textit{IB} 100) & \textbf{45.4} & 45.2 & 45.1\\
Places-LT & \textbf{38.8} & 38.5 & 38.6\\
ImageNet-LT & \textbf{53.0} & 47.0 & 52.1\\
iNaturalist 2018 & \textbf{70.0} & 58.3 & 69.8 \\
\bottomrule
\end{tabular}
}
\label{tab:ablation}
\vspace{0mm}
\end{table}
\begin{table*}[t]
\footnotesize
\centering
\caption{Top-1 accuracy over all classes on test time shifted CIFAR-100-LT with imbalance ratio of 50.}
\vspace{0px}
{
	\begin{tabular}{l|ccccc|c|ccccc}
    \toprule
    Dataset & \multicolumn{5}{c|}{Forward} & Uniform & \multicolumn{5}{c}{Backward}\\
    \midrule
    Imbalance ratio & 50 & 25 & 10 & 5 & 2 & 1 & 2 & 5 & 10 & 25 & 50 \\
    \midrule
	Causal Norm & 63.7 & 61.6 & 58.7 & 55.9 & 51.5 & 48.1 & 44.7 & 41.2 & 38.3 & 35.6 & 33.6 \\
	Balanced Softmax & 59.6 & 58.5 & 56.9 & 54.8 & 52.2 & 49.9 & 47.5 & 45.1 & 42.7 & 40.9 & 39.9 \\
	Softmax & 65.9 & 63.4 & 59.7 & 55.6 & 50.1 & 45.5 & 40.8 & 35.2 & 30.5 & 26.8 & 23.9 \\
    \midrule
	PC Causal Norm & 66.1 & 62.9 & 58.8 & 55.6 & 51.2 & 48.1 & 45.7 & 44.2 & 43.4 & 44.3 & 44.9 \\
	PC Balanced Softmax & 65.9 & 63.1 & 59.5 & \textbf{56.3} & 52.2 & 49.9 & 47.9 & 46.9 & 46.4 & 47.3 & 48.4 \\
	PC Softmax & 66.0 & 63.2 & 59.2 & 55.9 & 52.4 & 49.5 & 47.5 & 46.7 & 46.2 & 47.4 & 49.0 \\
    \midrule
	\textbf{LADE} & \textbf{67.4} & \textbf{64.7} & \textbf{60.2} & \textbf{56.3} & \textbf{52.8} & \textbf{50.5} & \textbf{48.2} & \textbf{47.4} & \textbf{46.6} & \textbf{48.1} & \textbf{49.4} \\
	\bottomrule
	\end{tabular}
}
\label{tab:cifar100_imb0.02_shift}
\vspace{0mm}
\end{table*}

\cite{choi2020regularized} introduces $\lambda$ to control the instability induced from directly using the DV representation.
The model suffers a severe performance drop on ImageNet-LT and iNaturalist 2018 when the regularizer term for DV representation is not used ($\lambda = 0$).
$\alpha$ represents the regularization strength of LADER on logits, as mentioned in Section 4.4.
Without LADER ($\alpha = 0$), performance degradation is observed, demonstrating the efficacy of LADER.

\subsection{Additional results on variously shifted test label distributions}
In Section 4.3, we show that our LADE achieves state-of-the-art performance on variously shifted test label distribution with ImageNet-LT, which is the large-scale long-tailed dataset.
We further conduct experiments on the small-scale dataset, CIFAR-100-LT, to ensure the consistent effectiveness of our LADE loss.
For the training set, we use CIFAR-100-LT with an imbalance ratio of 50.
The shifted test set is constructed by the same setting in Section 4.3.
As shown in Table~\ref{tab:cifar100_imb0.02_shift}, LADE outperforms all the other methods, which is consistent with the results on ImageNet-LT (Table 6).
We can also reconfirm the effectiveness of the PC strategy.
These results from CIFAR-100-LT and ImageNet-LT imply that our PC strategy and LADE work well on both small-scale and large-scale datasets.

\subsection{Additional confidence calibration results}
We report the additional results of LADE against other methods in the perspective of confidence calibration, using the same datasets from the section above, CIFAR-100-LT with an imbalance ratio of 50 for the small-scale dataset and ImageNet-LT for the large-scale dataset.
Following~\cite{ovadia2019can, kull2019beyond}, we estimate the quality of calibration on two datasets with four metrics:
\begin{itemize}[leftmargin=*]
\item \textbf{Expected Calibration Error}
\begin{align}
\text{ECE}=\frac{1}{N} \sum _{m=1} ^ {M} |B_m| \cdot |acc(B_m) - conf(B_m)|,
\end{align}
\item \textbf{Classwise Expected Calibration Error}
\begin{equation}
\begin{aligned}
&\text{Classwise-ECE} \\ &=\frac{1}{C} \sum_{j=1}^{C} \sum_{m=1} ^ {M} |B_{m, j}| \cdot |acc(B_{m, j}) - conf(B_{m, j})|
\end{aligned}
\end{equation}

\item \textbf{Brier Score}
\begin{align}\text{Brier} = \sum _{i=1} ^ {N} \sum_{c=1} ^ {C} (p(y_i=c | x_i; \theta) - \mathbbm{1}(y_i = c))^2,
\end{align}
\item \textbf{Negative Log Likelihood}
\begin{align}
\text{NLL} = -\sum _{i=1} ^ {N} \log p(y_i | x_i; \theta),
\end{align}
\end{itemize}
where $N$ is the total number of test samples $(x_i, y_i)$,
$C$ is the total number of classes,
$M(=20)$ is the total number of bins,
each bin $B_m$ is the set of indices of test samples where $\frac{m-1}{M} < p(y_i|x_i; \theta) \leq \frac{m}{M}$,
$|B_m|$ is the total number of samples inside the bin $B_m$,
$acc(B_m) = \frac{1}{|B_m|} \sum _{i \in B_m} \mathbbm{1}(\arg\max_{y_j} p(y_j|x_i;\theta) = y_i)$,
and $conf(B_m) = \frac{1}{|B_m|} \sum _{i \in B_m} p(y_i|x_i;\theta)$.
The bin $B_{m, j}$ is the set of indices of test samples where the class for the samples is $j$, and the other definitions $|B_{m, j}|$, $acc(B_{m, j})$ and $conf(B_{m, j})$ are exactly same as the above.

\begin{table*}[h]
\footnotesize
\centering
\caption{Confidence calibration results on CIFAR-100-LT with imbalance ratio of 50.}
{
\begin{tabular}{l c c c c c}
\toprule
Method & Accuracy & ECE & Classwise ECE & Brier & NLL \\
\midrule
Causal Norm                    & 48.1 & 0.150 &0.00483& 0.689 & 2.13 \\
Balanced Softmax               & 49.9 & 0.168 &0.00461& 0.673 & 2.07 \\
Softmax                        & 45.5 & 0.249 &0.00680& 0.769 & 2.50 \\
\midrule
PC Softmax                     & 49.5 & 0.174 &0.00472& 0.678 & 2.10 \\
LADE                           & \textbf{50.5} &\textbf{0.148}&\textbf{0.00434}& \textbf{0.658} & \textbf{2.02} \\
\bottomrule
\end{tabular}
}
\label{tab:09_confidence}
\end{table*}
\begin{table*}[h]
\footnotesize
\centering
\caption{Confidence calibration results on ImageNet-LT.}
{
\begin{tabular}{l c c c c c}
\toprule
Method & Accuracy & ECE & Classwise ECE & Brier & NLL \\
\midrule
Causal Norm                    & 52.0 & 0.108 &0.000461 & 0.634 & 2.42 \\
Balanced Softmax               & 52.1 & 0.061 &0.000406& 0.621 & 2.20 \\
Softmax                        & 48.2 & 0.140 &0.000603& 0.688 & 2.47 \\
\midrule
PC Softmax                     & 52.8 & {0.057} &0.000411& 0.615 & \textbf{2.17} \\
LADE                           & \textbf{53.0} & \textbf{0.035} & \textbf{0.000406} & \textbf{0.611} & 2.18 \\
\bottomrule
\end{tabular}
}
\label{tab:10_confidence}
\end{table*}
Table~\ref{tab:09_confidence} and~\ref{tab:10_confidence} summarize the calibration results on CIFAR-100-LT and ImageNet-LT datasets, respectively.
For all the evaluation metrics, LADE shows better overall calibration results than baseline methods.
These observations demonstrate that our proposed LADE is effective in terms of calibration on both small-scale (CIFAR-100-LT) and large-scale (ImageNet-LT) datasets.

\clearpage
\begin{figure*}[t]
\begin{center}
\includegraphics[width=\textwidth]{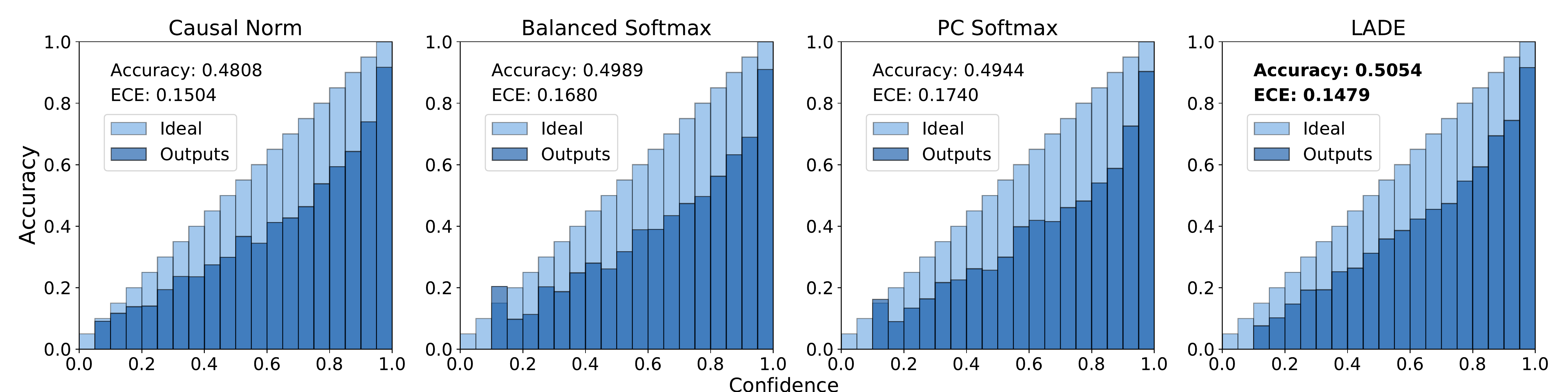}
\end{center}
\vspace{-10px}
\caption{Reliability diagrams of ResNet-32~\cite{he2016deep} on CIFAR-100-LT with imbalance ratio of 50.
}
\label{fig:5_cifar_confidence_calibration}
\end{figure*}

\end{document}